\documentclass[runningheads]{llncs}
\usepackage{graphicx}

\usepackage{tikz}
\usepackage{comment}
\usepackage{amsmath,amssymb} 
\usepackage{color}

\usepackage{epsfig}
\usepackage{subfig}
\usepackage{booktabs}
\usepackage{amsfonts}
\usepackage{enumerate}

\newcommand{\tabincell}[2]{\begin{tabular}{@{}#1@{}}#2\end{tabular}}

\begin{document}
\pagestyle{headings}
\mainmatter
\def\ECCVSubNumber{2460}  

\title{Learning Feature Embeddings \\ for Discriminant Model based Tracking} 

\titlerunning{Learning Feature Embeddings for Discriminant Model based Tracking}
%

\author{Linyu Zheng\inst{1,2} \and
Ming Tang\inst{1,3} \and
Yingying Chen\inst{1,2,4} \and
Jinqiao Wang\inst{1,2,4} \and
Hanqing Lu\inst{1,2}}
\authorrunning{L. Zheng et al.}
%
\institute{National Laboratory of Pattern Recognition, Institute of Automation, \\
  Chinese Academy of Sciences, Beijing 100190, China \and
School of Artificial Intelligence, University of Chinese Academy of Sciences, \\
  Beijing 100049, China \and
Shenzhen Infinova Limited \and
ObjectEye Inc. \\
\email{\{linyu.zheng, tangm, yingying.chen, jqwang, luhq\}@nlpr.ia.ac.cn}}

\maketitle

\begin{abstract}
After observing that the features used in most online discriminatively trained trackers are not optimal, in this paper, we propose a novel and effective architecture to learn optimal feature embeddings for online discriminative tracking. Our method, called DCFST, integrates the solver of a discriminant model that is differentiable and has a closed-form solution into convolutional neural networks. Then, the resulting network can be trained in an end-to-end way, obtaining optimal feature embeddings for the discriminant model-based tracker. As an instance, we apply the popular ridge regression model in this work to demonstrate the power of DCFST. Extensive experiments on six public benchmarks, OTB2015, NFS, GOT10k, TrackingNet, VOT2018, and VOT2019, show that our approach is efficient and generalizes well to class-agnostic target objects in online tracking, thus achieves state-of-the-art accuracy, while running beyond the real-time speed. Code will be made available.
\end{abstract}

\section{Introduction}

Visual object tracking is one of the fundamental problems in computer vision. Given the initial state of a target object in the first frame, the goal of tracking is to estimate the states of the target in the subsequent frames~\cite{OTB2015,GPRT,GPAS}. Despite the significant progress in recent years, visual tracking remains challenging because the tracker has to learn a robust appearance model from very limited online training samples to resist many extremely challenging interferences, such as large appearance variation and heavy background clutters. In general, the key problem of visual tracking is how to construct a tracker which can not only tolerate the appearance variation of the target, but also exclude the background interference, while keeping the running speed that is as high as possible.

There has been significant progress in deep convolutional neural networks (CNNs) based trackers in recent years. From a technical standpoint, existing state-of-the-art CNNs-based trackers mainly fall into two categories. (1) The one is to treat tracking as a problem of similarity learning and is only trained offline. Typically, SINT~\cite{SINT}, SiamFC~\cite{SiamFC}, and SiamRPN~\cite{SiamRPN} belong to this category. Although these trackers achieve state-of-the-art performance on many challenging benchmarks, the lack of online learning prevents them from integrating background in an online and adaptive way to improve their discriminative power. Therefore, they are severely affected by heavy background clutters, hindering the further improvement of localization accuracy. (2) The other is to apply CNNs features to the trackers which are discriminatively trained online. Most of these trackers, such as HCF~\cite{HCF}, ECO~\cite{ECO}, LSART~\cite{LSART}, and fdKCF*~\cite{fdKCF}, extract features via the CNNs which are trained on ImageNet~\cite{ImageNet} for object classification task. Obviously, these features are not optimal for the visual tracking task. Therefore, such trackers are not able to make the visual tracking task sufficiently benefit from the powerful ability of CNNs in feature embedding learning. Even though CFNet~\cite{CFNet} and CFCF~\cite{CFCF} learnt feature embeddings for online discriminatively trained correlation filters-based trackers by integrating the KCF solver~\cite{KCF} into the training of CNNs, the negative boundary effect~\cite{CFLB} in KCF severely degrades the quality of the feature embeddings they learn as well as their localization accuracy. Therefore, it is hard for their architectures to achieve high accuracy in online tracking.

To solve the above problem, in this paper, we propose a novel and effective architecture to learn optimal feature embeddings for online discriminative tracking. Our proposed network receives a pair of images, training image and test image, as its input in offline training~\footnote{In this paper, offline training refers to training deep convolutional neural networks, that is the process of learning feature embeddings, whereas discriminative training refers to training discriminant models, such as ridge regression and SVM. In our approach, each iteration of the offline training involves discriminative training.}. First, an efficient sub-network is designed to extract the features of real and dense samples around the target object from each input image. Then, a discriminant model that is differentiable and has a closed-form solution is trained to fit the samples in the training image to their labels. Finally, the trained discriminant model predicts the labels of samples in the test image, and the predicted loss is calculated. In this way, the discriminant model is trained without circulant and synthetic samples like in KCF, avoiding the negative boundary effect naturally. On the other hand, because it is differentiable and has a closed-form solution, its solver can be integrated into CNNs as a layer with forward and backward processes during training. Therefore, the resulting network can be trained in an end-to-end way, obtaining optimal feature embeddings for the discriminant model-based tracker.

As an instance, we apply the popular ridge regression model in this work to demonstrate the power of the proposed architecture because ridge regression model not only is differentiable and has a closed-form solution, but also has been successfully applied by many modern online discriminatively trained trackers~\cite{KCF,SRDCF,BACF,ECO,LSART,fdKCF} due to its simplicity, efficiency, and effectiveness. In particular, we employ Woodbury identity~\cite{Woodbury} to ensure the efficient training of ridge regression model when high-dimensional features are used because it allows us to address the dependence of time complexity on the dimension of features. Moreover, we observed that the extreme imbalance of foreground-background samples encountered during network training slows down the convergence speed considerably and severely reduces the generalization ability of the learned feature embeddings if the commonly used mean square error loss is employed. In order to address this problem, we modify the original shrinkage loss~\cite{Shrinkage} which is designed for deep regression learning and apply it to achieve efficient and effective training.

In online tracking, given the position and size of a target object in the current frame, we extract the features of real and dense samples around the target via the above trained network, \emph{i.e.}, learned feature embeddings, and then train a ridge regression model with them. Finally, in the next frame, the target is first located by the trained model, and then its position and size are refined by ATOM~\cite{ATOM}.

It is worth mentioning that the core parts of our approach are easy-to-implement in a few lines of code using the popular deep learning packages. Extensive experiments on six public benchmarks, OTB2015, NFS, GOT10k, TrackingNet, VOT2018, and VOT2019, show that the proposed tracker DCFST, \emph{i.e.}, learning feature embeddings with Differentiable and Closed-Form Solver for Tracking, is efficient and generalizes well to class-agnostic target objects in online tracking, thus achieves state-of-the-art accuracy on all six datasets, while running beyond the real-time speed. Fig.~\ref{fig:SpeedAccuracy} provides a glance of the comparison between DCFST and other state-of-the-art trackers on OTB2015. We hope that our simple and effective DCFST will serve as a solid baseline and help ease future research in visual tracking.


\section{Related Work}

In this section, we briefly introduce recent state-of-the-art trackers, with a special focus on the Siamese network based ones and the online discriminatively trained ones. In addition, we also shortly describe the recent advances in meta-learning for few-shot learning since our approach shares similar insights to theirs.

\subsection{Siamese Network Based Trackers}

Recently, Siamese network based trackers~\cite{SINT,SiamFC,SiamDCN} have received much attention for their well-balanced accuracy and speed. These trackers treat visual tracking as a problem of similarity learning. By comparing the target image patch with the candidate patches in a search region, they consider the candidate with the highest similarity score as the target object. A notable characteristic of such trackers is that they do not perform online learning and update, achieving high FPS in online tracking. Typically, SiamFC~\cite{SiamFC} employed a fully-convolutional Siamese network to extract features and then used a simple cross-correlation layer to evaluate the candidates in a search region in a dense and efficient sliding-window way. GCT~\cite{SiamGCN} introduced the graph convolution into SiamFC. SiamRPN~\cite{SiamRPN} enhanced the accuracy of SiamFC by adding a region proposal sub-network after the Siamese network. CRPN~\cite{CSiamRPN} improved the accuracy of SiamRPN using a cascade structure. SiamDW~\cite{SiamRPN} and SiamRPN++~\cite{SiamRPN++} enabled SiamRPN to benefit from deeper and wider networks. Even though these trackers achieve state-of-the-art performance on many benchmarks, a key limitation they share is their inability to integrate background in an online and adaptive way to improve their discriminative power. Therefore, they are severely affected by heavy background clutters, hindering the further improvement of localization accuracy.

Different from Siamese trackers, our DCFST can integrate background in an online and adaptive way through training discriminant models. Therefore, it is more robust to background clutters than Siamese trackers.

\subsection{Online Discriminatively Trained Trackers}

In contrast to Siamese network based trackers, another family of trackers~\cite{KCF,SRDCF,fdKCF} train discriminant models online to distinguish the target object from its background. These trackers can effectively utilize the background of the target, achieving impressive discriminative power on multiple challenging benchmarks. The latest such trackers mainly focus on how to take advantage of CNNs features effectively. HCF~\cite{HCF}, ECO~\cite{ECO}, fdKCF*~\cite{fdKCF}, and LSART~\cite{LSART} extracted features via the CNNs which are trained on ImageNet for object classification task and applied them to online discriminatively trained trackers. Obviously, these features are not optimal for the visual tracking task. Therefore, these trackers are not able to make the visual tracking task sufficiently benefit from the powerful ability of CNNs in feature embedding learning. In order to learn feature embeddings for online discriminatively trained correlation filters-based trackers, CFNet~\cite{CFNet} and CFCF~\cite{CFCF} integrated the KCF~\cite{KCF} solver into the training of CNNs. However, it is well known that KCF has to resort to circulant and synthetic samples to achieve the fast training of the filters, introducing the negative boundary effect and degrading the localization accuracy of trackers. Even though CFNet relaxed the boundary effect by cropping the trained filters, its experimental results show that this heuristic idea produces very little improvement. CFCF employed CCOT tracker~\cite{CCOT}, that is less affected by the boundary effect, in online tracking. However, its offline training does not aim to learn feature embeddings for CCOT, but for KCF, and its running speed is far away from real-time due to the low efficiency of CCOT. Therefore, we argue that it is hard for their architectures to achieve high accuracy and efficiency simultaneously in online tracking.

Different from CFNet and CFCF, our DCFST shares similar insights to DiMP's~\cite{DiMP}. Both DCFST and DiMP propose an end-to-end trainable tracking architecture, capable of learning feature embeddings for online discriminatively trained trackers without circulant and synthetic samples. In addition, the training of their discriminant models in both offline training and online tracking are identical. Therefore, they do not suffer from the negative boundary effect and can track target objects more accurately than CFNet and CFCF. The main differences between our DCFST and DiMP are as follows. (1) The architecture of DiMP forces it to use the square-shaped fragments to approximate the real foreground samples, ignoring the actual aspect ratio of the target object. In contrast, that of DCFST is more flexible, allowing us to sample identically to the actual size of the target object. Therefore, the foreground samples are approximate in DiMP, but relatively accurate in DCFST. (2) DiMP designed an iterative method to train its discriminant model, which cannot always guarantee an optimal solution. Whereas, DCFST uses a close-form solver which can always guarantee an optimal solution. (3) From the perspective of implementation, DCFST is much simpler than DiMP. The components and codes of the core parts of DCFST are much fewer than those of DiMP. Experiments show that DCFST outperforms CFNet and DiMP in tracking accuracy with large margins and it also outperforms CFCF in both tracking accuracy and speed with large margins.

\begin{figure}[t]
\centering
\includegraphics[height=46mm,width=120mm]{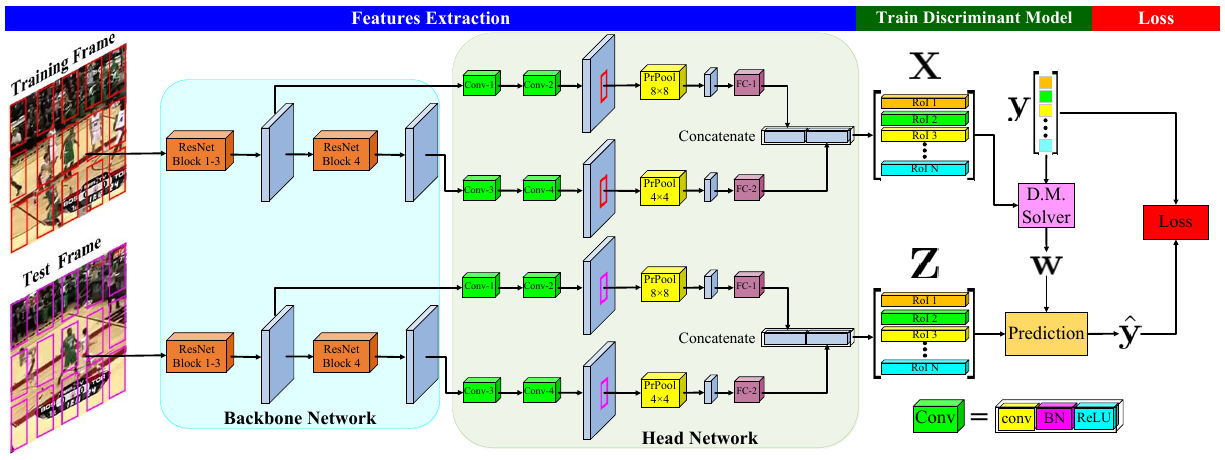}
\caption{Full architecture of the proposed network for learning feature embeddings. For each input image, $N$ regions of interest (RoIs) with the target size are generated by uniform sampling. ResNet Block-3 and Block-4 backbone feature maps extracted from the input image are first passed through two convolutional layers to obtain two learned feature maps. Fixed-size feature maps of each RoI are then extracted using PrPool layers and further mapped to feature vectors using fully-connected layers. $\mathbf{X}$ and $\mathbf{Z}$ are data matrices composed of the learned feature vectors of all training and test samples, respectively. A discriminant model $\mathbf{w}$ is trained to fit the samples in $\mathbf{X}$ to their labels. Finally, $\mathbf{w}$ predicts the labels of the samples in $\mathbf{Z}$, and the predicted loss is calculated. Best viewed in color.}
\label{fig:framework}
\end{figure}

\subsection{Meta-Learning Based Few-Shot Learning}

Meta-learning studies what aspects of the learner effect generalization across a distribution of tasks. Recently, differentiable convex optimization based meta-learning approaches greatly promote the development of few-shot learning. Instead of the nearest-neighbor based learners, MetaOptNet~\cite{MetaOptNet} used discriminatively trained linear predictors as base learners to learn representations for few-shot learning, and it aimed at learning feature embeddings that generalize well to novel categories under a linear classification rule. Bertinetto \emph{et al.}~\cite{DCFS} proposed both closed-form and iterative solvers, based on ridge regression and logistic regression components, to teach a CNN to use standard machine learning tools as part of its internal model, enabling it to adapt to novel data quickly.

To our best knowledge, the proposed DCFST is the first tracker to integrate the solver of a discriminant model that is differentiable and has a closed-form solution into the training of CNNs to learn optimal feature embeddings for online discriminative tracking without circulant and approximate samples. Experiments on multiple challenging benchmarks show that our approach achieves state-of-the-art accuracy at beyond the real-time speed and also sets a simple yet strong baseline for visual tracking. Therefore, we believe that it would promote the development of high-accuracy and real-time tracking.

\section{Learning Feature Embeddings}

The main task of an online discriminatively trained tracker is to train a discriminant model $\mathbf{w}$ which is able to not only fit the training samples well online, but also generalize well to the test samples. It is well known that not only different modeling methods, such as nearest neighbor and ridge regression, directly effect the generalization ability of $\mathbf{w}$, but features are also crucial to it. Therefore, our approach, DCFST, is developed by designing an architecture to learn optimal feature embeddings for discriminant model-based trackers, rather than more powerful discriminant models as most modern online discriminatively trained trackers did, to improve tracking accuracy.

As shown in Fig.~\ref{fig:framework}, the proposed network receives a pair of $288\times 288$ RGB images, training image and test image, as its input in offline training. It consists of the following three parts: features extraction network, discriminant model solver, and loss function. This section will present them one by one.

\subsection{Features Extraction Network}
\label{sec:FEN}

For each input image, the features extraction consists of the following five steps.

\begin{enumerate} [(1)]
\item $N$ RoIs with the target size are generated by uniform sampling across the whole image. In addition, the vector $\mathbf{y}\in \mathbb{R}^{N\times 1}$ containing their Gaussian labels is constructed as done in KCF~\cite{KCF} with standard deviation of $0.25$.
\item ResNet~\cite{ResNet} Block-3 and Block-4 backbone feature maps are extracted from the input image and then passed through two convolutional layers to obtain two learned feature maps. Their strides are $8\times 8$ and $16 \times 16$, respectively. Here, all convolutional kernels are $3\times 3$ and all convolutional layers are followed by BatchNorm~\cite{BN} and ReLU.
\item Fixed-size feature maps of each RoI are respectively extracted from the above two learned feature maps using PrPool layers~\cite{IoUNet} and further mapped to feature vectors using fully-connected layers. Specifically, the output sizes of two PrPool layers are $8\times 8$ and $4\times 4$, respectively, and both following fully-connected layers output a $512$-dimensional feature vector.
\item Two $512$-dimensional feature vectors of each RoI are concatenated to produce the learned feature vector of it. Its dimension, denoted as $D$, is $1024$.
\item The learned feature vectors of all training RoIs form the training data matrix $\mathbf{X}\in \mathbb{R}^{N\times D}$. The test data matrix $\mathbf{Z}\in \mathbb{R}^{N\times D}$ is obtained in the same way.
\end{enumerate}

It is worth noting that different from CFCF and CFNet whose training data matrices are circulant and most training samples are virtual ones, and different from DiMP whose training and test samples are always assumed to be square, in our DCFST, training data matrix is non-circulant and all training and test samples are real sampled identically to the actual size of the target object.

\subsection{Discriminant Model Solver}
\label{sec:RR}

We train a discriminant model that is differentiable and has a closed-form solution to fit the samples in $\mathbf{X}$ to their labels by integrating its solver into the proposed network. Because the discriminant model is differentiable and has a closed-form solution, its solver can be integrated into CNNs as a layer with forward and backward processes during training. As an instance, we apply the popular ridge regression model in this work to demonstrate the power of the proposed architecture. Ridge regression model has been confirmed to be simple, efficient and effective in the field of visual object tracking~\cite{BACF,ECO,LSART,MKCFup,fdKCF}. It can not only exploit all foreground and background samples to train a good regressor, but also effectively use high-dimensional features as the risk of over-fitting can be controlled by l2-norm regularization. Most importantly, it is differentiable and has a closed-form solution.

The optimization problem of ridge regression can be formulated as
\begin{equation}
\label{eq:RR}
\min_\mathbf{w}\left \| \mathbf{X}\mathbf{w}-\mathbf{y} \right \|_2^{2}+\lambda \left \| \mathbf{w} \right \|_2^{2},
\end{equation}
where $\mathbf{X}\in \mathbb{R}^{N\times D}$ contains $D$-dimensional feature vectors of $N$ training samples, $\mathbf{y}\in \mathbb{R}^{N\times 1}$ is the vector containing their Gaussian labels, and $\lambda>0$ is the regularization parameter. Its optimal solution can be expressed as
\begin{equation}
\label{eq:w}
\mathbf{w}^{*} = \left ( \mathbf{X}^{\top}\mathbf{X}+\lambda \mathbf{I}\right )^{-1}\mathbf{X}^{\top}\mathbf{y}.
\end{equation}

Solving for $\mathbf{w}^{*}$ by directly using Eq.~\ref{eq:w} is time-consuming because $\mathbf{X}^{\top}\mathbf{X} \in \mathbb{R}^{D\times D}$ and the time complexity of matrix inversion is $O\left ( D^3 \right )$. Even if we obtain $\mathbf{w}^{*}$ by solving a system of linear equations with Gaussian elimination method, the time complexity is still $O\left ( D^3/2 \right )$, hindering the efficient training of the model when high-dimensional features are used. To address the dependence of the time complexity on the dimension of features, we employ the Woodbury formula~\cite{Woodbury}
\begin{equation}
\label{eq:Woodbury}
\left ( \mathbf{X}^{\top}\mathbf{X}+\lambda \mathbf{I}\right )^{-1}\mathbf{X}^{\top}\mathbf{y} = \mathbf{X}^{\top}\left ( \mathbf{X}\mathbf{X}^{\top}+\lambda \mathbf{I}\right )^{-1}\mathbf{y},
\end{equation}
where $\mathbf{X}\mathbf{X}^{\top}\in \mathbb{R}^{N\times N}$. It is easy to see that the right hand of Eq.~\ref{eq:Woodbury} allows us to solve for $\mathbf{w}^{*}$ in time complexity $O\left ( N^3/2 \right )$. Usually, the number of online training samples is smaller than the dimension of features in tracking, \emph{i.e}., $N<D$. Therefore, in order to solve for $\mathbf{w}^{*}$ efficiently, we use the right hand of Eq.~\ref{eq:Woodbury} if the dimension of the learned feature vectors is larger than the number of training samples, \emph{i.e}, $D>N$. Otherwise, the left hand is used.

Last but not least, when we integrate the ridge regression solver into the training of CNNs, it is necessary to calculate $\partial \mathbf{w}^{*}/\partial \mathbf{X}$ in the backward process. Fortunately, $\partial \mathbf{w}^{*}/\partial \mathbf{X}$ is easy to be automatically obtained using the popular deep learning packages, such as TensorFlow and PyTorch, where automatic differentiation is supported. Specifically, during network training, given $\mathbf{X}$ and $\mathbf{y}$, only one line of code is necessary to solve for $\mathbf{w}^{*}$ in the forward process and there is no code needed in the backward process.

\subsection{Fast Convergence with Shrinkage Loss}

There exists extreme imbalance between foreground and background samples during network training. This problem slows down the convergence speed considerably and severely reduces the generalization ability of the learned feature embeddings if the commonly used mean square error loss $\mathcal{L}_{mse} = \left \| \mathbf{y}-\hat{\mathbf{y}} \right \|_2^{2}$ is employed, where $\mathbf{y}$ and $\hat{\mathbf{y}}=\mathbf{Z}\mathbf{w}^{*}$ are vectors containing the ground-truth labels and the predicted labels of the $N$ test samples in $\mathbf{Z}$, respectively. In fact, this is because most background samples are easy ones and only a few hard samples provide useful supervision, making the network training difficult.

To address this problem and make the network training efficient and effective, we propose a new shrinkage loss
\begin{equation}
\label{eq:shrinkage}
\mathcal{L}=\left \| \frac{\exp\left ( \mathbf{y} \right ) \odot \left ( \mathbf{y}-\hat{\mathbf{y}} \right )}{1+\exp\left (  a\cdot \left ( c-\left | \mathbf{y}-\hat{\mathbf{y}} \right | \right ) \right )} \right \|_2^{2},
\end{equation}
where $a$ and $c$ are hyper-parameters controlling the shrinkage speed and location, respectively, and the absolute value and the fraction are element-wise. Specifically, $\mathcal{L}$ down-weights the losses assigned to easy samples and mainly focuses on a few hard ones, preventing the vast number of easy backgrounds from overwhelming the learning of feature embeddings. It is easy-to-implement in a few lines of code using the current deep learning packages.

In fact, Eq.~\ref{eq:shrinkage} is a modified version of the original shrinkage loss~\cite{Shrinkage}
\begin{equation}
\label{eq:shrinkage_original}
\mathcal{L}_{o}=\left \| \frac{\exp\left ( \mathbf{y} \right ) \odot \left ( \mathbf{y}-\hat{\mathbf{y}} \right )}{1+\exp\left (  a\cdot \left ( c-\left ( \mathbf{y}-\hat{\mathbf{y}} \right ) \right ) \right )} \right \|_2^{2}.
\end{equation}
Mathematically, the main difference between Eq.~\ref{eq:shrinkage} and Eq.~\ref{eq:shrinkage_original} is that a sample is regarded as an easy one if its predicted value is larger than its label and their difference is less than $c$ in Eq.~\ref{eq:shrinkage_original}, whereas in Eq.~\ref{eq:shrinkage}, we only consider the absolute difference to determine whether a sample is easy or not. In our experiments, we found that this modification can not only accelerate the convergence speed and reduce the validation loss in offline training, but also improve the accuracy in online tracking. Therefore, it may provide other researchers a better choice.

Moreover, it is worth noting that the motivations for using shrinkage loss in our approach and \cite{Shrinkage} are quite different. In \cite{Shrinkage}, the purpose of using shrinkage loss is to prevent the discriminant model, \emph{i.e.}, filters, from under-fitting to a few foreground samples after iterative training. However, there is no such concern in our approach because the discriminant model is trained directly by a close-form solution rather than an iterative method. The purpose of using shrinkage loss in our approach is similar to that of using focal loss~\cite{Focal} in training detectors, that is, preventing the vast number of easy backgrounds from overwhelming the learning of feature embeddings.

Based on the above design and discussions, the resulting network can be trained in an end-to-end way, obtaining optimal feature embeddings for the discriminant model-based tracker. Ideally, in online tracking, the learned feature embeddings should make the corresponding discriminant model, \emph{e.g.}, ridge regression model in this work, trained with the features extracted via them robust not only to the large appearance variation of the target object, but also to the heavy background clutters, thereby improving tracking accuracy.

\section{Online Tracking with Learned Feature Embeddings}

\subsection{Features Extraction}

Suppose $\left ( \mathbf{p}_t,\mathbf{s}_t \right )$ denotes the position and size of the target object in frame $t$. Given frame $t$ and $\left ( \mathbf{p}_t,\mathbf{s}_t \right )$, we sample a square patch centered at $\mathbf{p}_t$, with an area of $5^2$ times the target area, and then resize it to $288\times 288$. Finally, the training data matrix $\mathbf{X}_t$ is obtained using the approach presented in Sec.~\ref{sec:FEN}. Similarly, given frame $t+1$ and $\left ( \mathbf{p}_t,\mathbf{s}_t \right )$, the test data matrix $\mathbf{Z}_{t+1}$ can be obtained.

\subsection{Online Learning and Update}

In online tracking, according to Sec.~\ref{sec:RR}, we can train a ridge regression model using the right hand or the left hand of Eq.~\ref{eq:Woodbury} for online discriminative tracking. However, the time complexities of both sides are cubical with respect to $D$ or $N$, hindering real-time performance. In order to solve for $\mathbf{w}^{*}$ more efficiently, we adopt the Gauss-Seidel based iterative approach~\cite{SRDCF}. Specifically, taking the left hand of Eq.~\ref{eq:Woodbury} as an example, given $\mathbf{X}_t$ and $\mathbf{w}^{*}_{t-1}$, we decompose $\mathbf{X}_{t}^{\top}\mathbf{X}_{t}+\lambda \mathbf{I}$ into a lower triangular $\mathbf{L}_t$ and a strictly upper triangular $\mathbf{U}_t$, \emph{i.e.}, $\mathbf{X}_{t}^{\top}\mathbf{X}_{t}+\lambda \mathbf{I}=\mathbf{L}_t+\mathbf{U}_t$. Then, $\mathbf{w}_{t}^{*}$ can be efficiently solved by iterative expressions:
\begin{subequations}
\begin{align}
\label{eq:Gauss-Seidel-a}
&\mathbf{w}^{*(j)}_{t}\leftarrow \mathbf{w}^{*}_{t-1}, & j=0,\\
\label{eq:Gauss-Seidel-b}
&\mathbf{w}^{*(j)}_{t}\leftarrow \mathbf{L}_t \setminus \left ( \mathbf{X}_{t}^{\top}\mathbf{y}-\mathbf{U}_t \mathbf{w}^{*(j-1)}_{t} \right ), & j>0,
\end{align}
\label{eq:Gauss-Seidel}
\end{subequations}
where $\mathbf{w}^{*}_{t-1}$ is the trained model at frame $t-1$, and $j$ indicates the number of iterations. In practice, $5$ iterations are enough for the satisfactory $\mathbf{w}^{*}_{t}$. Note that this iterative method is efficient because Eq.~\ref{eq:Gauss-Seidel-b} can be solved efficiently with forward substitution, and the time complexity of each iteration is $O\left ( D^2 \right )$ instead of $O\left ( D^3/2 \right )$.

In order to locate the target object robustly, updating the appearance model is necessary. Following the popular updating method used in~\cite{KCF,BACF,CFNet,fdKCF}, we update $\mathbf{X}_t$ by means of the linear weighting approach, \emph{i.e.},
\begin{equation}
\label{eq:update}
\begin{aligned}
&\widetilde{\mathbf{X}}_1 = \mathbf{X}_1, \\
&\widetilde{\mathbf{X}}_t = \left ( 1-\delta  \right )\widetilde{\mathbf{X}}_{t-1}+\delta \mathbf{X}_t, \qquad t>1,
\end{aligned}
\end{equation}
where $\delta$ is the learning rate. As a result, instead of $\mathbf{X}_t$, $\widetilde{\mathbf{X}}_t$ is used in Eq.~\ref{eq:Gauss-Seidel} for solving for $\mathbf{w}^{*}_{t}$. In addition, we keep the weight of $\mathbf{X}_{1}$ not being less than $0.25$ during updating because the initial target information is always reliable.

\subsection{Localization and Refine}

Given the trained model $\mathbf{w}^{*}_{t}$ and test data matrix $\mathbf{Z}_{t+1}$, we locate the target by $\hat{\mathbf{y}}_{t+1} = \mathbf{Z}_{t+1} \mathbf{w}^{*}_{t}$, and the sample corresponding to the maximum element of $\hat{\mathbf{y}}_{t+1}$ is regarded as the target object. After locating the target, we refine its bounding box by ATOM~\cite{ATOM} for more accurate tracking, similar to DiMP~\cite{DiMP}.

\section{Experiments}

Our DCFST is implemented in Python using PyTorch. On a single TITAN X(Pascal) GPU, employing ResNet-18 and ResNet-50 respectively as the backbone network, our DCFST-18 and DCFST-50 achieve tracking speeds of 35 FPS and 25 FPS, respectively. Code will be made available.

\subsection{Implementation Details}

{\noindent \bfseries Training Data.} To increase the generalization ability of the learned feature embeddings, we use the training splits of recent large-scale tracking datasets, including TrackingNet~\cite{TrackingNet}, GOT10k~\cite{GOT10k}, and LaSOT~\cite{LaSOT}, in offline training. During network training, each pair of training and test images is sampled from a video snippet within the nearest $50$ frames. For training image, we sample a square patch centered at the target, with an area of $5^2$ times the target area. For test image, we sample a similar patch, with a random translation and scale relative to the target's. These cropped patches are then resized to a fixed size $288\times 288$. We use image flipping and color jittering for data augmentation.

{\noindent \bfseries Training Setting.} We use the pre-trained model on ImageNet to initialize the weights of our backbone network and freeze them during network training. The weights of our head network are randomly initialized with zero-mean Gaussian distributions. We train the head network for 40 epochs with 1.5k iterations per epoch and 48 pairs of images per batch, giving a total training time of 30(40) hours for DCFST-18(50) on a single GPU. The ADAM~\cite{ADAM} optimizer is employed with initial learning rate of 0.005, using a factor $0.2$ decay every 15 epochs.

{\noindent \bfseries Parameters.} We sample $961$ RoIs in each input image, \emph{e.g.}, $N=961$. The regularization parameter $\lambda$ in Eq.~\ref{eq:RR} is set to 0.1. Two hyper-parameters $a$ and $c$ in Eq.~\ref{eq:shrinkage} are set to $10$ and $0.2$, respectively. The learning rate $\delta$ in Eq.~\ref{eq:update} is $0.01$.

\begin{table}[!t]
\caption{Ablation studies on OTB2015. (a) The mean AUC scores of different loss functions. Our modified shrinkage loss achieves the best result. (b) The mean AUC scores and FPSs of different $N$s. $N=961$ achieves good balance.}
\begin{center}
\subfloat[Loss Function.]
{\resizebox{0.5\columnwidth}{!}{
\begin{tabular}{c@{~~}c@{~~}c@{~~}c@{~~}}
\toprule[1pt]
Loss Function & \tabincell{c}{Mean \\Square Error} & \tabincell{c}{Original \\Shrinkage} & \tabincell{c}{Modified \\Shrinkage} \\ \midrule[0.5pt]
   Mean AUC   &                0.682               &                 0.698               &                0.709 \\
\bottomrule[1pt]
\end{tabular}}
\label{table:loss}}
\subfloat[Number of Samples.]
{\resizebox{0.47\columnwidth}{!}{
\begin{tabular}{c@{~~}c@{~~}c@{~~}c@{~~}c}
\toprule[1pt]
  $N$    &  361 ($19^2$)  &  625 ($25^2$)  &  961 ($31^2$) & 1369 ($37^2$) \\
\noalign{\smallskip} \midrule[0.5pt] \noalign{\smallskip}
Mean AUC &      0.692     &      0.701     &      0.709    &     0.707 \\
Mean FPS &       46       &        40      &        35     &       32   \\
\bottomrule[1pt]
\end{tabular}}
\label{table:samples}}
\end{center}
\end{table}

\subsection{Ablation Studies}

In this section, we will investigate the effect of choosing the loss function and hyper-parameter in our approach. Our ablation studies are based on DCFST-18 and performed on OTB2015~\cite{OTB2015}.

Table~\ref{table:loss} shows the mean AUC~\cite{OTB2015} scores of DCFST-18 with various loss functions. It can be seen that the tracking accuracy can be obviously improved ($1.6\%$ in AUC score) by using the original shrinkage loss instead of the mean square error one. This demonstrates that relaxing the imbalance of foreground-background samples in our approach is necessary and our choice of loss function is valid. Moreover, compared to the original shrinkage loss, our modified one can further improve the tracking accuracy by a large margin ($1.1\%$ in AUC score). This confirms the effectiveness of our improvement.

Table~\ref{table:samples} shows the mean AUC scores and FPSs of DCFST-18 with various $N$s. It can be seen that $N=961$ can not only provide beyond the real-time speed, but also high tracking accuracy. It is worth mentioning that the stride of ResNet Block-3 is $8$ pixel, and when $N=31^2$ and the size of input image is $288\times 288$, the interval between two adjacent samples is $7.68$ pixel. Therefore, the tracking accuracy will not be improved significantly when $N>961$.

\begin{figure}[!t]
\centering
\subfloat[]
{\includegraphics[height=45mm,width=53mm]{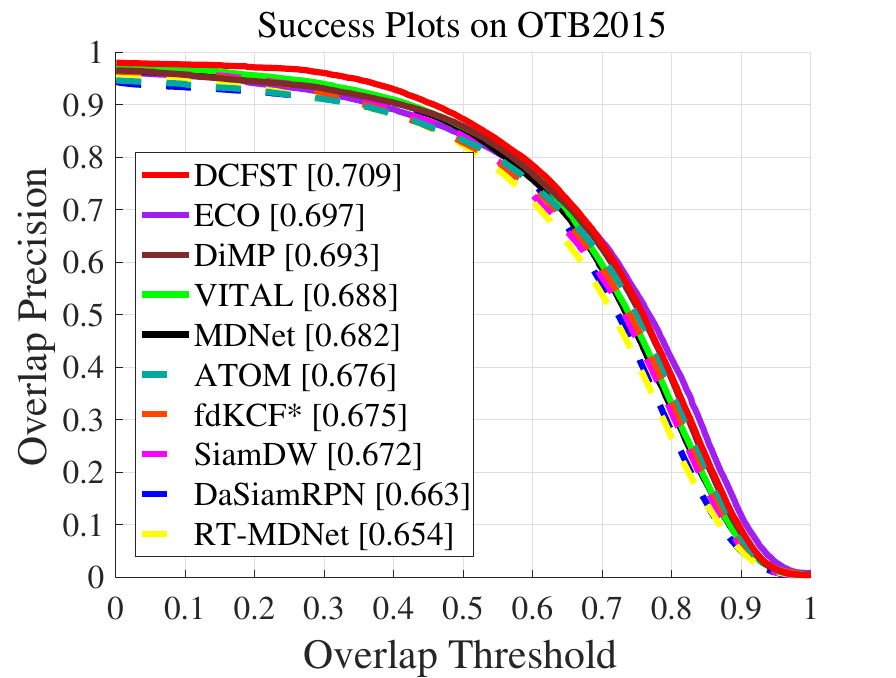}
\label{fig:OTB2015}}
\subfloat[]
{\includegraphics[height=45mm,width=63mm]{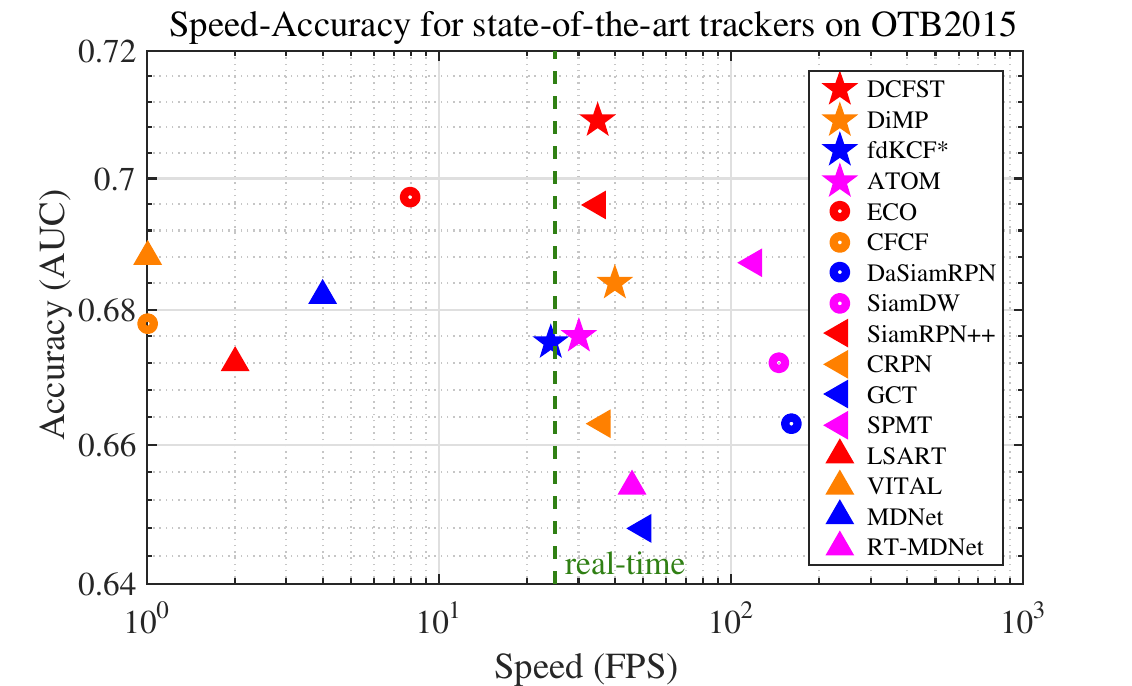}
\label{fig:SpeedAccuracy}}
\caption{State-of-the-art comparison on OTB2015. (a) The mean success plots of DCFST and nine state-of-the-art trackers. The mean AUC scores are reported in the legend. (b) Speed and accuracy plot of 16 state-of-the-art trackers. Our DCFST achieves the best accuracy, while running beyond the real-time speed.}
\end{figure}

\begin{table}[!t]
\caption{State-of-the-art comparison on OTB2015 in terms of mean overlap. The best three results are shown in red, blue, and magenta. Our DCFST outperforms its baseline tracker ATOM with a gain of $4\%$, using the same backbone network.}
\label{table:OTB2015}
\begin{center}
\resizebox{0.99\columnwidth}{!}{
\begin{tabular}{c@{~~}c@{~~}c@{~~}c@{~~}c@{~~}c@{~~}c@{~~}c@{~~}c@{~~}c@{~~}c}
\toprule[1pt]
Tracker      &         DCFST       &          DiMP        & fdKCF* & ATOM & SiamDW &           DaSiamRPN         & VITAL &  ECO  & MDNet & RT-MDNet\\ \midrule[0.5pt]
Mean Overlap & {\color{red} 0.872} & {\color{blue} 0.859} & 0.828  & 0.832& 0.840  &   {\color{magenta} 0.858}   & 0.857 & 0.842 & 0.852 & 0.822\\
\bottomrule[1pt]
\end{tabular}}
\end{center}
\end{table}

\begin{table}[!t]
\caption{State-of-the-art comparison on NFS in terms of mean AUC score. DiMP-18 and DiMP-50 are the DiMP tracker with ResNet-18 and ResNet-50 backbone network respectively. Both versions of our DCFST outperform other trackers.}
\label{table:NFS}
\begin{center}
\resizebox{0.99\columnwidth}{!}{
\begin{tabular}{c@{~~}c@{~~}c@{~~}c@{~~}c@{~~}c@{~~}c@{~~}c@{~~}c@{~~}c@{~~}c@{~~}c}
\toprule[1pt]
Tracker  &        DCFST-50     &        DCFST-18      &        DiMP-50          & DiMP-18 & ATOM  &  UPDT  &  ECO  & CCOT  & MDNet &  HDT & SFC \\ \midrule[0.5pt]
Mean AUC & {\color{red} 0.641} & {\color{blue} 0.634} & {\color{magenta} 0.620} &  0.610  & 0.584 &  0.537 & 0.466 & 0.488 & 0.429 & 0.403 & 0.401\\
\bottomrule[1pt]
\end{tabular}}
\end{center}
\end{table}

\begin{table}[!t]
\caption{State-of-the-art comparison on the GOT10k test set in terms of average overlap (AO), and success rates (SR) at overlap thresholds 0.5 and 0.75. DaSiamRPN-18 is the DaSiamRPN tracker with ResNet-18 backbone network. Our DCFST-50 outperforms other trackers with large margins.}
\label{table:GOT10k}
\begin{center}
\resizebox{0.99\columnwidth}{!}{
\begin{tabular}{c@{~~}c@{~~}c@{~~}c@{~~}c@{~~}c@{~~}c@{~~}c@{~~}c@{~~}c}
\toprule[1pt]
Tracker  &        DCFST-50     &          DCFST-18       &         DiMP-50      & DiMP-18 & ATOM  & DaSiamRPN-18 & CFNet & SiamFC & GOTURN  \\ \midrule[0.5pt]
  AO     & {\color{red} 0.638} & {\color{magenta} 0.610} & {\color{blue} 0.611} &  0.579  & 0.556 &    0.483     & 0.374 & 0.348  & 0.347  \\
SR(0.50) & {\color{red} 0.753} & {\color{magenta} 0.716} & {\color{blue} 0.717} &  0.672  & 0.634 &    0.581     & 0.404 & 0.353  & 0.375  \\ \
SR(0.75) & {\color{red} 0.498} & {\color{magenta} 0.463} & {\color{blue} 0.492} &  0.446  & 0.402 &    0.270     & 0.144 & 0.098  & 0.124  \\
\bottomrule[1pt]		
\end{tabular}}
\end{center}
\end{table}

\begin{table}[!t]
\caption{State-of-the-art comparison on the TrackingNet test set in terms of precision, normalized precision, and success. Our DCFST-50 achieves top results.}
\label{table:TrackingNet}
\begin{center}
\resizebox{0.99\columnwidth}{!}{
\begin{tabular}{c@{~~}c@{~~}c@{~~}c@{~~}c@{~~}c@{~~}c@{~~}c@{~~}c@{~~}c@{~~}c}
\toprule[1pt]
Tracker       &        DCFST-50     &         DCFST-18        &         DiMP-50         & DiMP-18  & ATOM  &          SiamRPN++        &  CRPN  &  SPMT  & DaSiamRPN & CFNet \\ \midrule[0.5pt]
Precision     & {\color{red} 0.700} &           0.682         & {\color{magenta} 0.687} &  0.666   & 0.648 &   {\color{blue} 0.694}    & 0.619  &  0.661 &   0.591   & 0.533 \\
Norm. Prec.   & {\color{red} 0.809} &           0.797         &   {\color{blue} 0.801}  &  0.785   & 0.771 & {\color{magenta} 0.800}   & 0.746  &  0.778 &   0.733   & 0.654 \\
Success (AUC) & {\color{red} 0.752} & {\color{magenta} 0.739} &  {\color{blue} 0.740}   &  0.723   & 0.703 &          0.733            & 0.669  &  0.712 &   0.638   & 0.578 \\
\bottomrule[1pt]		
\end{tabular}}
\end{center}
\end{table}

\begin{figure}[!t]
\centering
\includegraphics[height=33mm,width=39mm]{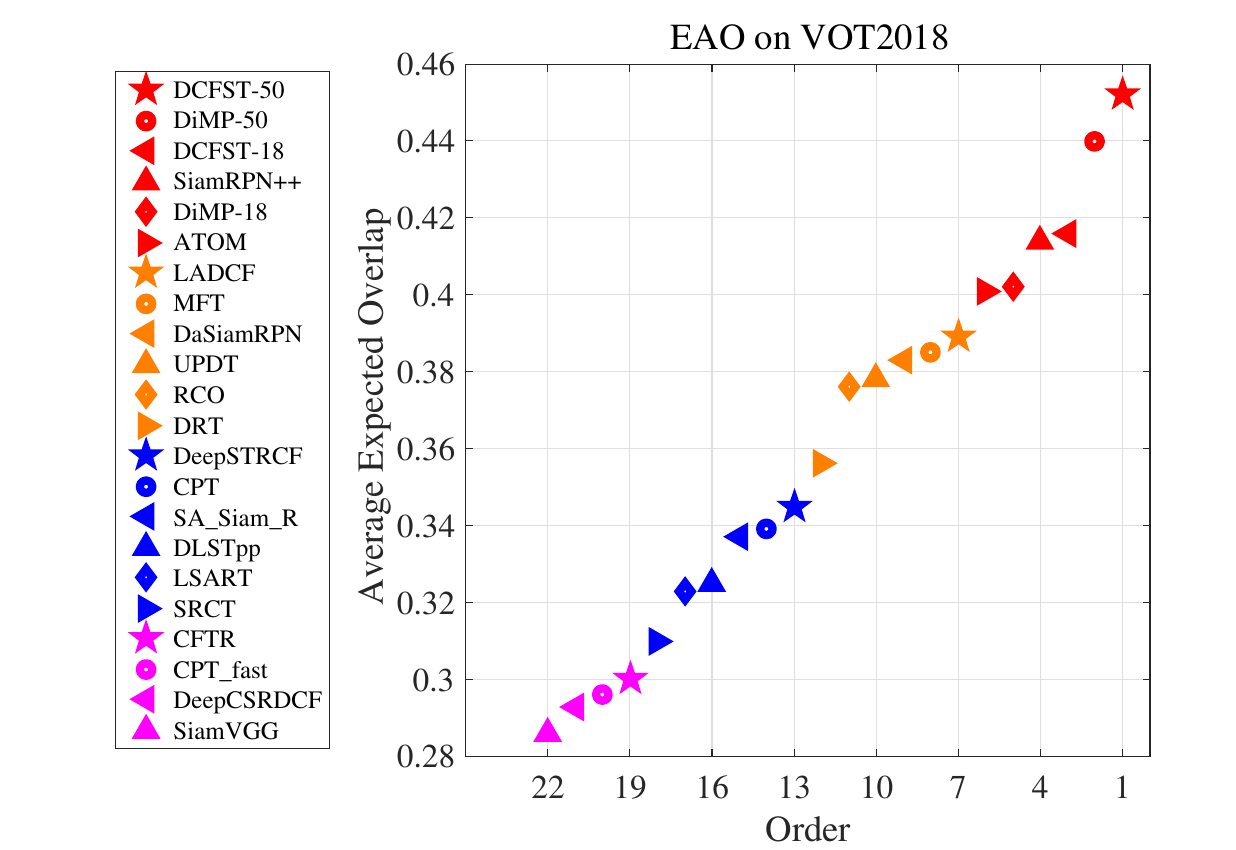}
\includegraphics[height=33mm,width=39mm]{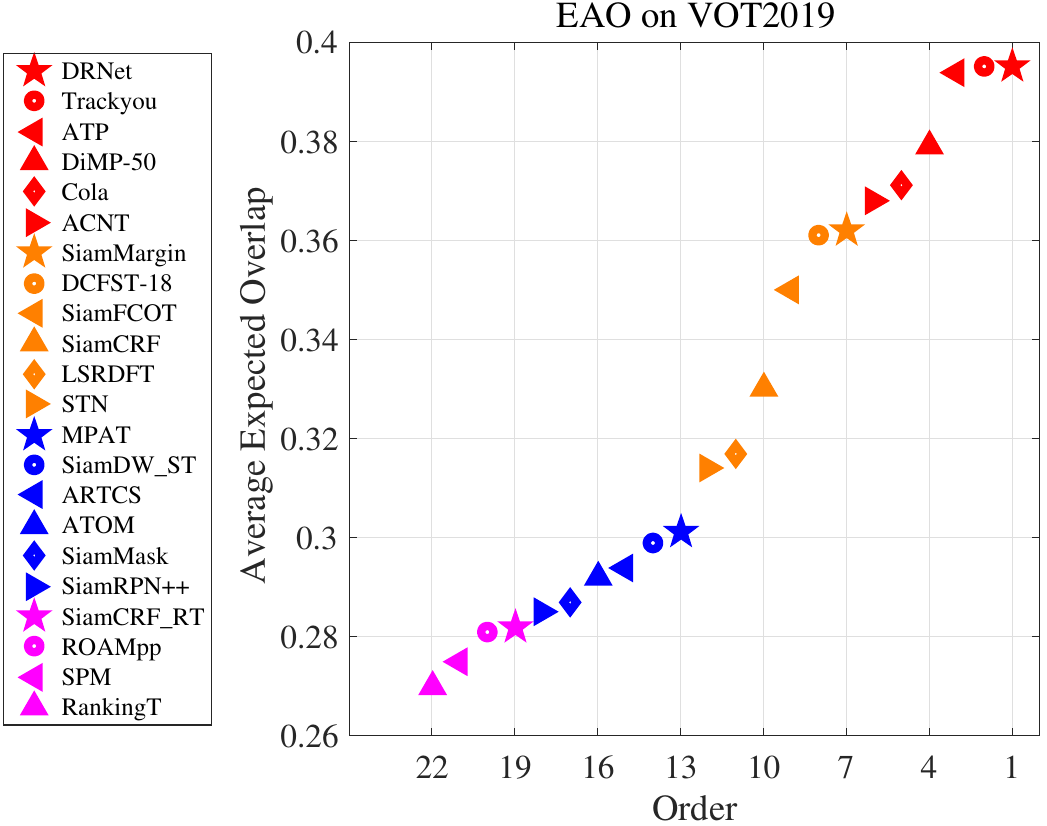}
\includegraphics[height=33mm,width=39mm]{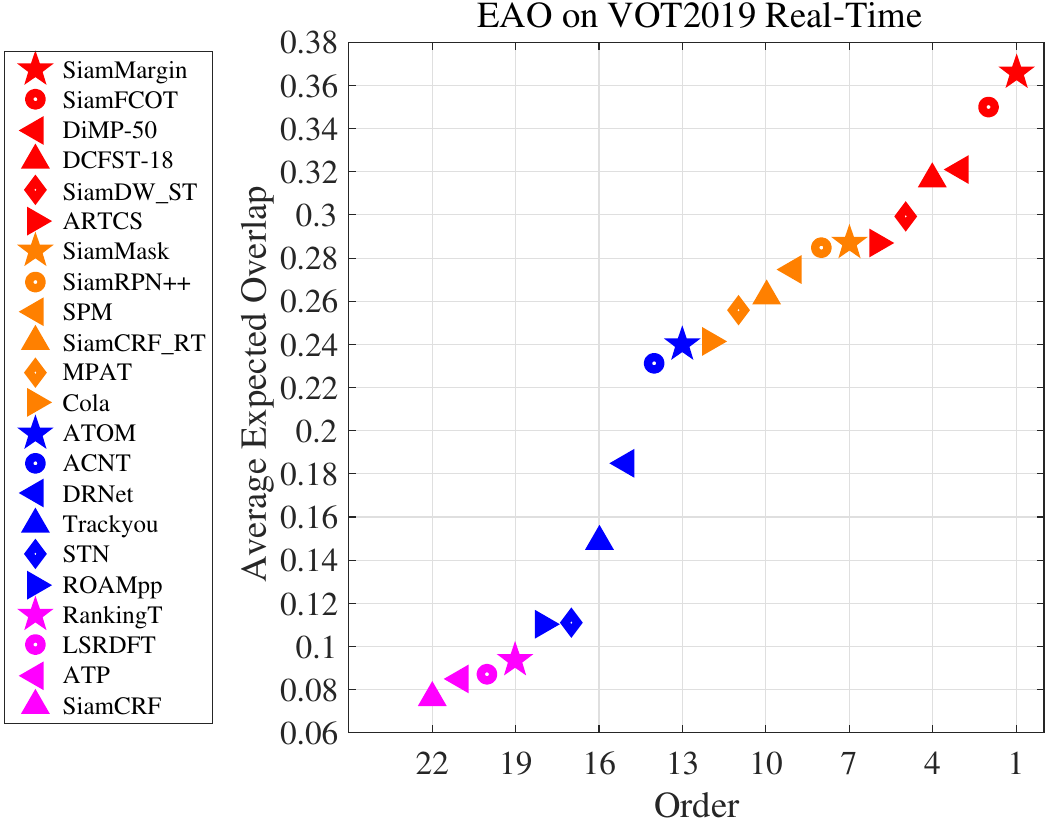}
\caption{State-of-the-art comparison on VOT2018, VOT2019, and VOT2019 real-time challenges in terms of expected average overlap (EAO).}
\label{fig:VOT}
\end{figure}

\subsection{State-of-the-art Comparisons}

{\noindent \bfseries OTB2015~\cite{OTB2015}.} OTB2015 is the most popular benchmark for the evaluation of trackers, containing 100 videos with various challenges. On the OTB2015 experiment, we first compare our DCFST-18 against nine state-of-the-art trackers, DiMP~\cite{DiMP}, fdKCF*~\cite{fdKCF}, ATOM~\cite{ATOM}, SiamDW~\cite{SiamRPN+}, DaSiamRPN~\cite{DaSiamRPN}, VITAL~\cite{VITAL}, ECO~\cite{ECO}, MDNet~\cite{MDNet}, and RT-MDNet~\cite{RTMDNet}. Success plot, mean overlap precision and AUC score~\cite{OTB2015} are employed to quantitatively evaluate all trackers. Fig.~\ref{fig:OTB2015} and Table~\ref{table:OTB2015} show the results. Our DCFST-18 obtains the mean AUC score and overlap precision of $70.9\%$ and $87.2\%$, outperforming the second best trackers (ECO and DiMP) with significant gains of $1.2\%$ and $1.3\%$, respectively. Additionally, Fig.~\ref{fig:SpeedAccuracy} shows the comparison of DCFST-18 with the above trackers along with SiamRPN++~\cite{SiamRPN++}, CRPN~\cite{CSiamRPN}, GCT~\cite{GCNT}, SPMT~\cite{SPMT}, LSART~\cite{LSART}, and CFCF~\cite{CFCF} in both mean AUC score and FPS. Our DCFST-18 achieves the best trade-off between accuracy and speed among all state-of-the-art trackers.

{\noindent \bfseries NFS~\cite{NFS}.} We evaluate our DCFST on the 30 FPS version of NFS benchmark which contains 100 challenging videos with fast-moving objects. On the NFS experiment, we compare DCFST against DiMP, ATOM, UPDT~\cite{UPDT}, ECO, CCOT~\cite{CCOT} along with the top-3 trackers, MDNet, HDT~\cite{HDT}, and SFC~\cite{SFC}, evaluated by NFS. All trackers are quantitatively evaluated by AUC score. Table~\ref{table:NFS} shows the results. Our DCFST-18 and DCFST-50 obtain the mean AUC scores of $63.4\%$ and $64.1\%$, outperforming the latest state-of-the-art trackers DiMP-18 and DiMP-50 with gains of $2.4\%$ and $2.1\%$, respectively. Additionally, DCFST-18 surpasses its baseline tracker ATOM~\footnote{We state the relationship between DCFST and ATOM in supplementary materials.} with a significant gain of $5.1\%$.

{\noindent \bfseries GOT10k~\cite{GOT10k}.} We evaluate our DCFST on the test set of GOT10k which is a large-scale tracking benchmark and contains over 9000 training videos and 180 test videos. Here, the generalization capabilities of the tracker to unseen object classes is of major importance. Therefore, to ensure a fair comparision, on the GOT10k experiment, we retrain our DCFST and then compare it against DiMP, ATOM, DaSiamRPN along with the top-3 trackers, CFNet~\cite{CFNet}, SiamFC~\cite{SiamFC}, and GOTURN~\cite{GOTURN}, evaluated by GOT10k. Following the GOT10k challenge protocol, all trackers are quantitatively evaluated by average overlap, and success rates at overlap thresholds 0.5 and 0.75. Table~\ref{table:GOT10k} shows the results. In terms of success rate at overlap thresholds 0.5, our DCFST-18 and DCFST-50 obtain the scores of $71.6\%$ and $75.3\%$, outperforming the latest state-of-the-art trackers DiMP-18 and DiMP-50 with gains of $4.4\%$ and $3.6\%$, respectively. Additionally, DCFST-18 surpasses its baseline tracker ATOM with a significant gain of $8.2\%$.

{\noindent \bfseries TrackingNet~\cite{TrackingNet}.} We evaluate our DCFST on the test set of TrackingNet which is a large-scale tracking benchmark and provides 511 test videos to assess trackers. On the TrackingNet experiment, we compare DCFST against seven state-of-the-art trackers, DiMP, ATOM, SiamRPN++, DaSiamRPN, SPMT, CRPN, and CFNet. Following the TrackingNet challenge protocol, all trackers are quantitatively evaluated by precision, normalized precision, and AUC score. Table~\ref{table:TrackingNet} shows the results. Our DCFST-18 and DCFST-50 obtain the mean AUC scores of $73.9\%$ and $75.2\%$, outperforming the latest state-of-the-art trackers DiMP-18 and DiMP-50 with gains of $1.6\%$ and $1.2\%$, respectively. Additionally, DCFST-18 surpasses its baseline tracker ATOM with a significant gain of $3.6\%$.

{\noindent \bfseries VOT2018/2019~\cite{VOT2018,VOT2019}.} We evaluate our DCFST on the 2018 and 2019 versions of Visual Object Tracking (VOT) challenge which contain 60 sequences, respectively. On the VOT2018 experiment, we compare DCFST against SiamRPN++, DiMP, ATOM along with the top-16 trackers on VOT2018 challenge. On the VOT2019 experiment, we compare DCFST-18 against the top-21 trackers on VOT2019 and VOT2019 real-time challenges, respectively. Following the VOT challenge protocol, all trackers are quantitatively evaluated by expected average overlap (EAO). Fig.~\ref{fig:VOT} shows the results. (1) On the VOT2018 challenge, our DCFST-18 and DCFST-50 obtain the EAO scores of $0.416$ and $0.452$, outperforming the latest state-of-the-art trackers DiMP-18 and DiMP-50 with gains of $1.4\%$ and $1.2\%$, respectively. Additionally, DCFST-50 outperforms all other state-of-the-art trackers including SiamRPN++, and DCFST-18 surpasses its baseline tracker ATOM with a significant gain of $1.5\%$. (2) On the VOT2019 and VOT2019 real-time challenges, our DCFST-18 achieves the EAO scores of $0.361$ and $0.317$, respectively, surpassing its baseline tracker ATOM with significant gains of $6.9\%$ and $7.7\%$, respectively. There are seven latest trackers perform well than DCFST-18 on VOT2019 challenge, however, five of them are obviously lower than DCFST-18 in terms of real-time performance. Although SiamRPN++ employs stronger backbone network (ResNet-50) and more training datas (ImageNet DET, ImageNet VID, YouTube, and COCO) than DCFST-18, DCFST-18 consistently outperforms it on all three challenges with large margins.

\section{Conclusion}

A novel and state-of-the-art tracker DCFST is proposed in this paper. By integrating the solver of a discriminant model that is differentiable and has a closed-form solution into convolutional neural networks, DCFST learns optimal feature embeddings for the discriminant model-based tracker in offline training, thus improves its accuracy and robustness in online tracking. Extensive experiments on multiple challenging benchmarks show that DCFST achieves state-of-the-art accuracy and beyond the real-time speed, and also sets a simple yet strong baseline for visual tracking due to its simplicity, efficiency and effectiveness.

\noindent\textbf{Acknowledgements.} This work was supported by the Research and Development Projects in the Key Areas of Guangdong Province (No. 2020B010165001). This work was also supported by National Natural Science Foundation of China under Grants 61772527, 61976210, 61806200, 61702510 and 61876086.

\clearpage
%
%
\bibliographystyle{splncs04}
\bibliography{egbib}
\end{document}